\pdfoutput=1

\documentclass[11pt]{article}
\usepackage{acl}
\usepackage{times}
\usepackage{latexsym}
\usepackage[T1]{fontenc}
\usepackage[utf8]{inputenc}
\usepackage{microtype}
\usepackage{inconsolata}
\usepackage{graphicx}
\usepackage{amsmath}
\usepackage{booktabs}
\usepackage{todonotes}
\usepackage{makecell}

\usepackage{todonotes}

\title{Enhancing Scene Transition Awareness in Video Generation via Post-Training}

\author{Hanwen Shen, Jiajie Lu, Yupeng Cao, Xiaonan Yang \\Stevens Institute of Technology \\}

\begin{document}

\maketitle

\begin{abstract}
Recent advances in AI-generated video have shown strong performance on \emph{text-to-video} tasks, particularly for short clips depicting a single scene. However, current models struggle to generate longer videos with coherent scene transitions, primarily because they cannot infer when a transition is needed from the prompt. Most open-source models are trained on datasets consisting of single-scene video clips, which limits their capacity to learn and respond to prompts requiring multiple scenes. Developing scene transition awareness is essential for multi-scene generation, as it allows models to identify and segment videos into distinct clips by accurately detecting transitions. To address this, we introduce the Transition-Aware Video (TAV) dataset with multi-scene clips and captions that explicitly state scene segmentation and transition structure. Our focus is on how prompt semantics and dataset annotations about temporal context affect text-to-video generation. Post-training on TAV improves alignment between the scene count implied by prompt and the scene count produced by the model, while preserving visual quality. 

\end{abstract}

\section{Introduction}\label{sec:intro}
The ability to generate visual content from natural language has rapidly improved in recent years, driven by the emergence of powerful generative models such as diffusion (e.g., \citet{ho2020ddpm}, \citet{song2021ddim}, \citet{rombach2022ldm}, \citet{Achiam2023GPT4}) and visual autoregressive model (e.g., \citet{oord2016pixelcnn}, \citet{kalchbrenner2017vpn}, \citet{chen2020imagegpt}, \citet{chen2023magvit}). These methods have become central to modern text-to-image and text-to-video systems, enabling high-quality results from simple prompts and forming the foundation of advanced T2V models such as \emph{Sora} \citep{brooks2024sora} and \emph{Kling} \citep{kuaishou2024kling}.

We observe that existing video generation models perform well on short clips with a single scene, but often struggle to maintain quality and coherence in longer, story-level videos. Open-source models like \emph{EasyAnimate} \citep{xu2024easyanimate} and \emph{CogVideo} \citep{hong2022cogvideo}, typically struggle to recognize the needs for scene transitions, i.e, they fail to generate the correct number of scenes as specified in the prompt, even when multiple distinct scenes are explicitly described. We evaluate the open-sourced models using 50 prompts that explicitly require the generation of two distinct scenes. As shown in Table \ref{tab:result}, the average number of scenes generated is approximately one, supporting our conclusion about the models' limited ability to handle multi-scene prompts.

One possible reason is that widely used video-text datasets, such as \emph{WebVid-10M} \citep{bain2022frozentimejointvideo}, \emph{Panda-70M} \citep{chen2024panda70mcaptioning70mvideos}, and \emph{MiraData} \citep{ju2024miradatalargescalevideodataset}, are largely composed of single-scene clips (over $90\%$), typically extracted using simple scene segmentation techniques. Thus, current models are rarely exposed to explicit scene transitions during training, which results in an out-of-distribution issue when a scene change is required at inference time. Given the strong generation capabilities of current models, we explore whether post-training them to recognize scene transitions in prompts can enhance overall performance, improving both coherence and visual quality.

\begin{table}[htpb!]
    \centering
    \begin{tabular}{ccc}
    \toprule 
    OpenSora & CogVideo & EasyAnimate \\
    \midrule
    1.12 & 1.48 & 1.22\\
    \bottomrule
    \end{tabular}
    \caption{Average number of scenes generated by models given prompts that explicitly indicate two scenes.}
    \label{tab:result}
\end{table}

Our contributions in this work includes:
\begin{itemize}
\item We design the \textbf{TAV} dataset to explicitly teach models how to handle scene transitions from prompt by post-training. The \textbf{TAV} dataset consists of pairs of 10-second video clips with scene transitions and their corresponding scene-wise descriptions. The clips are extracted from the \emph{Panda-70M} dataset, and for each clip, a large language model (LLM) is used to generate separate descriptions for each individual scene.

\item We conduct an experiment to compare the number of scenes generated by the original OpenSora model and the OpenSora model post-trained on the \textbf{TAV} dataset, using the same set of prompts. The results show that post-training with the \textbf{TAV} dataset increases the average number of scene, indicating improved understanding of scene transition requirements specified in the prompts. Notably, image quality remains unaffected, as measured by VBench \citep{huang2023vbenchcomprehensivebenchmarksuite}.
\end{itemize}

\section{Related Work}\label{sec:relatedwork}

\paragraph{Long video generation.} Increasing attention has been paid to generating long, story-driven videos in recent research. Early approaches leveraged GANs and VAEs to model video distributions, while models like VideoGPT \citep{yan2021videogpt} and TATS \citep{ge2022long} introduced discrete latent spaces and transformer-based architectures for improved temporal coherence. Transformer-based methods such as Phenaki \citep{villegas2022phenaki} further extended video length by generating token sequences conditioned on textual input. More recently, diffusion models have emerged as a powerful framework. Methods like LEO \citep{wang2023leo} and LVDM \citep{he2022lvdm} leverage hierarchical or latent motion spaces to synthesize long videos with enhanced continuity. NUWA-XL \citep{yin2023nuwa_xl} and GAIA-1 \citep{hu2023gaia1} adopt structured diffusion or world model approaches, while FreeNoise \citep{qiu2023freenoise} and Gen-L-Video \citep{wang2023gen_l_video} extend generation by aggregating noise-sampled or overlapping segments. StreamingT2V \citep{henschel2024streamingt2v} proposes an autoregressive framework with memory mechanisms to maintain appearance consistency over time.

\paragraph{Transition generation.} Scene transitions are essential for storytelling, enabling smooth shifts in time, space, or perspective. Traditional techniques such as fades, dissolves, wipes, and cuts are often implemented using predefined patterns, while morphing methods (\citet{wolberg1998image}, \citet{shechtman2010regenerative}) allow for smoother transitions by estimating pixel-level correspondences. Generative approaches like latent-space interpolation \citep{vandenOord2017neural} have been used to model semantic transitions, with applications in style transfer \citep{chen2018gated} and object transfiguration (\citet{sauer2022styleganxl},\citet{kang2023studiogan}). Recent advances explore data-driven methods for generative scene transitions. Seine \citep{Chen2023Seine} introduces a short-to-long video diffusion model focused on transitions and predictions. Loong \citep{wang2024loong} leverages autoregressive language models for minute-level multi-scene video generation, while VideoDirectorGPT \citep{lin2023videodirectorgpt} incorporates LLM-guided planning to ensure consistency across multiple scenes. Recent studies advanced long video generation by improving temporal alignment and scene coherence. 
TALC~\cite{Bansal2024TALC} and ShotAdapter~\cite{Kara2025ShotAdapter} introduced time-aligned captions and multi-shot control, 
while Mask$^2$DiT~\cite{Qi2025Mask2DiT}, MinT~\cite{Wu2024MinT}, and Presto~\cite{Yan2024LongVideoDiffusion} 
enhance diffusion transformers with masking and temporally aware attention. 
Mixture of Contexts~\cite{Cai2025Mixture} and MSG Score~\cite{Yoon2024MSGScore} 
further improve efficiency and evaluation for multi-scene long video synthesis. 

Incorporating specialized modules designed to improve consistency and capture scene transitions is essential for enhancing long video generation. From our perspective, evaluating the quality of the training dataset and identifying the most suitable and efficient data for this task should also be a top priority. To the best of our knowledge, this aspect has received limited attention, and our proposed \textbf{TAV} dataset aims to highlight its importance.

\paragraph{Datasets.}
Public video–text datasets can be roughly grouped by scale and focus.  
\emph{Web-scale corpora}—\textit{MiraData} (330 k long clips) \citep{ju2024miradata}, HD-\textsc{VILA}\,100M \citep{xue2022hdvila}, and auto-captioned sets like \textit{Panda-70M} \citep{chen2024panda70m} and \textit{InternVid} \citep{wang2023internvid}—supply hundreds of millions of paired frames that power minute-level diffusion/Transformer training. \emph{General short-video caption sets}, led by \textit{WebVid-10M} \citep{bain2022frozentimejointvideo} and \textit{HowTo100M} \citep{miech2019howto100m}, dominate text-to-video pre-training, while action-label datasets such as \textit{Kinetics-700} \citep{carreira2019kinetics700} and \textit{Moments-in-Time} \citep{monfort2019moments} emphasize clip-level semantics. Finally, a spectrum of \emph{domain-specific benchmarks} remains essential for evaluation and niche tasks: classical recognition/caption corpora (\textit{UCF101} \citep{soomro2012ucf101}, \textit{MSR-VTT} \citep{xu2016msrvtt}, \textit{ActivityNet-Captions} \citep{krishna2017activitynet}, \textit{YouCook2} \citep{zhou2018youcook2}); egocentric \textit{Ego4D} \citep{grauman2022ego4d}; face-centric \textit{CelebV-Text} \citep{yu2023celebvtext}; robotic \textit{BAIR} \citep{finn2017bair}; synthetic \textit{Moving MNIST} \citep{srivastava2015movingmnist}; and interpolation-oriented \textit{Vimeo-90K} \citep{xue2017vimeo90k}. Together, these resources span from thousands to hundreds of millions of videos, underpinning contemporary generative models across training, fine-tuning, and evaluation.

\section{Method}

In this section, we present the pipeline on preparing the \textbf{TAV} dataset.

\paragraph{Data source.} 
We first draw a sample of $500$ videos from the validation set of \emph{Panda-70M} dataset, which contains a total of $2,000$ videos. The sample was carefully constructed to ensure that its category distribution closely approximates that of the full dataset, effectively serving as a representative subset of the population. For the post-training stage, the sample is split into $480$ videos for training, $50$ for validation, and $50$ for testing.

\paragraph{Scene transition detection.} We modified the method in \emph{PySceneDetect}. Let $L(i,j)$ and $R(i,j)$ represent pixel values at position $(i,j)$ in two image frames for the same channel, and $N$ be the total number of pixels. We define the average pixel difference as:
$$
D(L, R)=\frac{1}{N}\sum_{i,j}|L(i,j)-R(i,j)|.
$$
Then we compute the average pixel difference in each HSV channel between consecutive frames, and define the overall frame change value as
\begin{align*}
V_t^{}=&w_H^{}\cdot D(H_t^{},H_{t-1}^{})\\
&+w_S^{}\cdot D(S_t^{},S_{t-1}^{})+w_V^{}\cdot D(V_t^{},V_{t-1}^{}).
\end{align*}
Here $w_H^{}, w_S^{}, w_V^{}$ are the weights assigned to each channel by the user. A scene cut is detected if $$V_t^{}>\mathrm{threshold}.$$

\paragraph{Scene transition extraction.} We apply the aforementioned scene transition detection method to the previously selected $500$ video samples. For each video, we retain only the first detected scene cut and extract a 10-second clip centered around it (combining 5 seconds before and 5 seconds after the transition point) to obtain a segment that contains a clear scene transition. If either side of the transition point does not contain a full 5 seconds of footage, we include as much as is available.

\paragraph{Video Data Caption.} After obtaining the 10-second clip containing two distinct scenes, we use \emph{BLIP} to generate separate textual descriptions for each scene. These descriptions are then combined into a single prompt that explicitly indicates a scene transition. For example: \emph{\{Previous scene: Superman is flying across the city; Next scene: He sees Batman fighting the Joker on a rooftop\}}. The \textbf{TAV} dataset consists of $500$ video–prompt pairs constructed in this manner.

\begin{table*}[bhtp!]
\centering
\begin{tabular}{lccccccc}
\toprule
& group & epoch & \makecell{average\\segments} & \makecell{aesthetic\\quality} & \makecell{overall\\consistency} & \makecell{dynamic\\degrees} & \makecell{imaging\\quality} \\
\midrule
Baseline & A & - & 1.180 & 0.510 & 0.045 & 0.203 & 0.652 \\
Baseline & B & - & 1.060 & 0.551 & 0.042 & 0.038 & 0.648 \\
Baseline & C & - & 1.120 & 0.517 & 0.049 & 0.089 & 0.643 \\
Post-trained & A & 16 & 1.840 & 0.401 & 0.060 & 0.783 & 0.575 \\
Post-trained & B & 16 & 1.800 & 0.405 & 0.062 & 0.789 & 0.592 \\
Post-trained & C & 16 & 1.740 & 0.395 & 0.062 & 0.816 & 0.584 \\
Post-trained & A & 24 & \textbf{2.380} & 0.436 & 0.052 & 0.538 & 0.647 \\
Post-trained & B & 24 & \textbf{2.700} & 0.419 & 0.054 & 0.526 & 0.630 \\
Post-trained & C & 24 & \textbf{2.900} & 0.429 & 0.060 & 0.517 & 0.599 \\
Post-trained & A & 36 & 2.300 & 0.430 & 0.053 & 0.643 & 0.608 \\
Post-trained & B & 36 & 2.520 & 0.425 & 0.054 & 0.643 & 0.622 \\
Post-trained & C & 36 & 2.400 & 0.443 & 0.057 & 0.515 & 0.616 \\
\bottomrule
ModelScope & - & - & - & 0.521 & 0.257 & 0.664 	&  0.586\\
VideoCrafter & - & - & - &	0.444 & 0.252	 & 0.897	&  0.572 \\
CogVideo & -& -& -&	0.382 & 0.077	& 0.422 &	 0.410 \\
LaVie & - & - & - &	0.549 & 0.264	 & 0.497	&  0.619 \\
\bottomrule
\end{tabular}
\caption{Evaluation metrics for baseline and post-trained models across different training epochs.}
\label{tab:epoch-post-train}
\end{table*}

\section{Experiments}

For consistency and brevity, we refer readers to Appendix A-C for implementation details, including code, and an example frame strip.

\paragraph{Implementation.} We fine-tune the \emph{OpenSora-Plan v1.3.1} \citep{lin2024opensoraplanopensourcelarge} model using the \textbf{TAV} dataset under a video-to-text generation setting. The training is performed using a single process with DeepSpeed Zero Stage 2 optimization. We utilize the \emph{google/mt5-xxl} text encoder and adopt a WFVAEModel (\emph{D8\_4x8x8}) pretrained from \emph{OpenSora-Plan v1.3.0} as the video autoencoder. The model processes 33-frame video clips at a resolution of 256$\times$256, with a sampling rate of 1 and frame rate of 8 FPS. Using a single H200 GPU, each training epoch completes in about 2 hours.

Key hyperparameters include a batch size of 1, 100 total training steps, a learning rate of $1 \times 10^{-5}$ with a constant scheduler, and \emph{bf16} mixed precision training. We use Exponential Moving Average (EMA) with a decay rate of 0.9999 starting from step 0. Gradient checkingpoint is enabled, and training is resumed from the latest checkpoint. Additional strategies include sparse 1D attention (with \emph{sparse\_n = 4}), temporal and spatial interpolation scales set to 1.0, and a guidance scale of 0.1. The model uses SNR-weighted loss (\emph{snr\_gamma = 5.0}) and adopts a \emph{v\_prediction} type for diffusion.


\paragraph{Experiment design.} To assess the model's performance, we construct three evaluation groups, each using a different version of the prompt to test the effectiveness of post-training.
\begin{itemize}
\item \textbf{Group A.}  This group uses prompts consisting of a single sentence without indicating any scene transition (e.g., \emph{\{Superman flying across the building\}}). It serves to demonstrate that the model is also capable of handling single-scene generation, highlighting its versatility beyond multi-scene transitions. 

\item \textbf{Group B.}  This group uses prompts containing two sentences that imply, but do not explicitly indicate, a scene transition. For example: \emph{\{Superman is flying across the building, and then sees Batman fighting the Joker on a rooftop\}}.

\item \textbf{Group C.} This group uses prompts that explicitly instruct a scene transition. For example: \emph{\{Previous scene: Superman is flying across the building; Next scene: Superman sees Batman fighting the Joker on a rooftop\}}.
\end{itemize}

We revise the prompts (originally generated from text descriptions of the $50$ test videos in the \textbf{TAV} dataset) into the three groups described above. These prompts are then applied to both the baseline and post-trained models to evaluate the average number of scene transitions and overall image quality. We show the results in Table \ref{tab:epoch-post-train}. 




\section{Result and Analysis}

As shown in Table \ref{tab:epoch-post-train}, the average number of scenes increases significantly after post-training. In the baseline model, particularly for Groups B and C, the average number of segments remains around 1, indicating a limited ability to recognize the need for multi-scene generation. In contrast, the post-trained model shows a substantial improvement, with the average number of segments even exceeding 2. These results demonstrate that post-training with the \textbf{TAV} dataset effectively enhances the model's capability for multi-scene generation.

Furthermore, post-training does not noticeably degrade video quality. \emph{Aesthetic Quality} and \emph{Motion Degree} have improved compared to ModelScope, LaVie, and CogVideo from data in VBench. On the contrary, it improves both \emph{dynamic consistency} and \emph{temporal smoothness}, enabling the model to generate more coherent motion and fluid scene transitions. As training progresses, we also observe gradual improvements in \emph{aesthetic quality} and \emph{imaging quality}, with metrics approaching or matching those of the baseline. These results suggest that post-training with the \textbf{TAV} dataset enhances multi-scene generation without compromising visual fidelity.

Moreover, even though post-training is conducted on a multi-scene dataset, the model still performs well on prompts requiring only a single scene (Group A). As shown in Table \ref{tab:epoch-post-train}, the post-trained model demonstrates strong performance not only when prompts explicitly indicate a two-scene structure (Group C), but also when the transition is only implicitly suggested (Group B).

With limited compute, we adopt OpenSora-Plan as the baseline. This lightweight model performs below average on scene-transition benchmarks. Our post-training substantially improves the baseline, but yields only modest gains over stronger baselines. Nevertheless, on \textbf{VBench} we observe higher \emph{Aesthetic Quality} and \emph{Motion Degree} than ModelScope, LaVie, and CogVideo.

\section{Conclusion}

In conclusion, our experiments demonstrate that prompt design plays a crucial role in controlling the number of scenes generated by T2V models. Furthermore, post-training on the \textbf{TAV} dataset significantly enhances the model’s ability to recognize and fulfill multi-scene generation requirements, especially when such intent is expressed explicitly in the prompt. Notably, we observe that, despite being trained on prompts with explicit scene-transition instructions, the post-trained model shows improved understanding and response to prompts that imply two scenes without explicitly stating the transition.

\section*{Limitation}
First, this study is a preliminary experiment. We only evaluate open-sourced state-of-the-art T2V models and use only a subset of videos from the Panda70M dataset due to limited computational resources.
Second, our experiments currently focus solely on multi-scene clips. A more comprehensive evaluation should include a mixture of both single-scene and multi-scene clips.
Third, it remains unclear which scene detection algorithm performs best in the T2V setting. The threshold configuration in our experiments is heuristically determined based on our prior experience.

\section*{Ethical Considerations}
\paragraph{Usage Rights.}Our Data is collected from Panda-70M, which is an open sourced dataset.  The Panda‑70M dataset is provided by Snap Inc.\ under a license for \emph{non-commercial, research purposes only}. Redistribution is permitted provided the original copyright notice, license terms, and disclaimers are retained.\cite{chen2024panda70m} Content are safe, covering diverse video domains, including animals, scenery, food, sports, activities, vehicles, tutorials, news and TV, gaming. Prompts are generated in English. 

Our main result is about the quantity of scene generated by T2V models, not about the content of scene, the risk of ethical concerns is minimal. All prompts are generated by open-sourced models. 

\section*{Acknowledgment }
We gratefully acknowledge the use of AI-assisted tools solely for grammatical corrections during manuscript preparation. No other aspects of the research— including conceptualization, experimental design, data analysis, or interpretation of results—were generated or modified by AI. All substantive content and conclusions were developed independently by the authors.

\section*{Implementation Details}

Code can be viewed at 
\href{https://github.com/hshen13/Enhancing-Scene-Transition-Awareness-in-Video-Generation-via-Post-Training}{github space}.

\bibliography{main}

\appendix
\onecolumn
\clearpage


\section*{Appendix A: Frame and Timeline Comparison}
\label{sec:frame_and_timeline_comparison}
\begin{figure*}[htp!]
    \centering
    \includegraphics[width=1\linewidth]{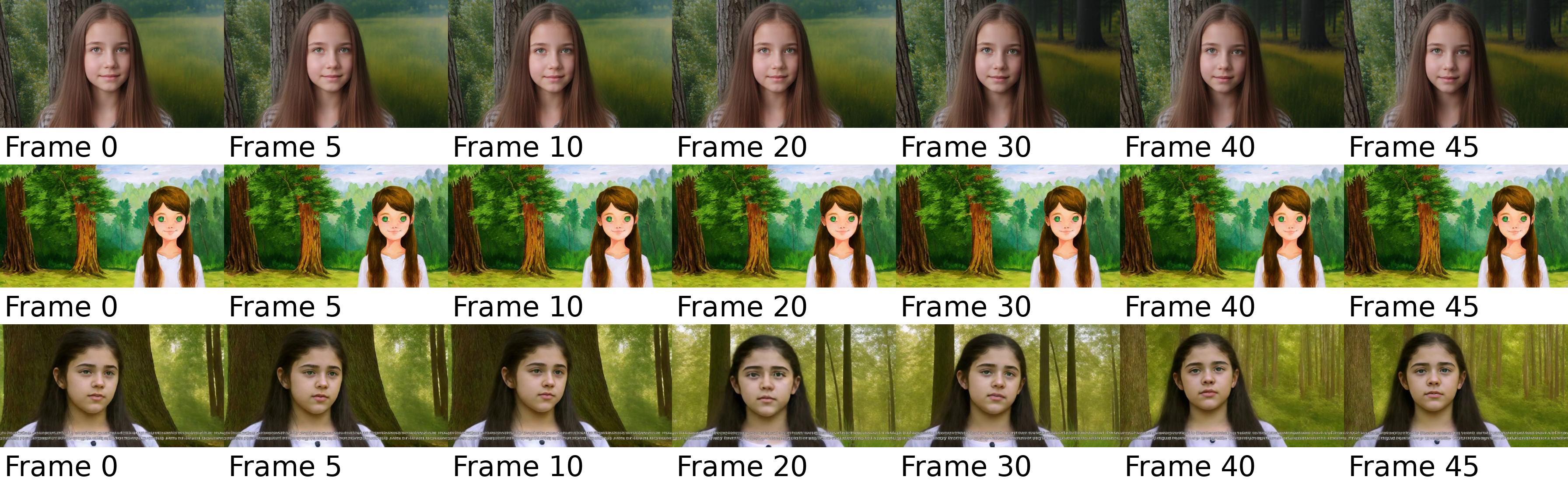}
    \caption{Frame–timeline comparison of three video generations. 
From \textbf{top} to \textbf{bottom}: (i) output generated by \textit{EasyAnimate}; 
(ii) output generated by \textit{OpenSora‑Plan} \emph{prior} to post‑training; 
and (iii) output generated by \textit{OpenSora‑Plan} \emph{after} 24 epochs of post‑training.  
The prompt used to generate is: \emph{``Previous scene: a girl with long hair and green eyes stands in front of a tree.  
Next scene: a painting of a forest with trees and grass''}}
    \label{fig:timeline_comparison_10}
\end{figure*}

\begin{figure*}[htp!]
    \centering
    \includegraphics[width=\linewidth]{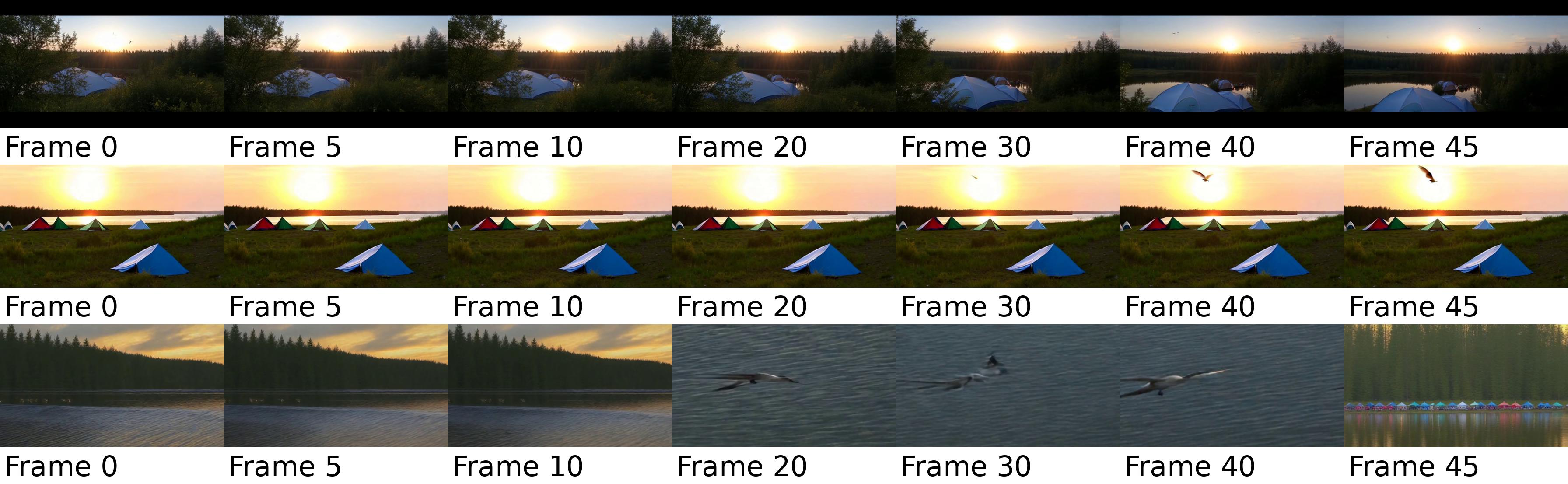}
    \caption{Frame–timeline comparison of three video generations. 
From \textbf{top} to \textbf{bottom}: (i) output generated by \textit{EasyAnimate}; 
(ii) output generated by \textit{OpenSora‑Plan} \emph{prior} to post‑training; 
and (iii) output generated by \textit{OpenSora‑Plan} \emph{after} 24 epochs of post‑training.  
The prompt used to generate is: \emph{``previous scene: a group of tents are set up in the woods; then next scene: a bird flying over the water at sunset
''}}
    \label{fig:timeline_comparison_10}
\end{figure*}

\clearpage
\section*{Appendix B: Prompt Examples}

This appendix provides a detailed listing of the prompt examples utilized in our experiments. Prompts are grouped into three categories: \textit{Single Scene Prompts}, \textit{Multi-Scene Prompts with Format}, and \textit{Multi-Scene Prompts without Format}. Out of a total of 50 prompts, 42 can be successfully rendered within the paper, whereas 8 fail to display correctly due to formatting or compatibility issues.

\subsection*{B.1 Single Scene Prompts}

\begin{enumerate}
  \item A man and woman sitting at a table on the beach.
  \item A group of tents are set up in the woods.
  \item A man and woman sitting at a table with drinks.
  \item A girl with long hair and green eyes stands in front of a tree.
  \item A boat is in the water near a rocky mountain.
  \item A yellow and black bird flying through a blue sky.
  \item A little girl in a wheelchair with a toy.
  \item A group of women holding signs in front of a crowd.
  \item A tall tower with a clock on top.
  \item A man in a suit and tie is talking to a woman.
  \item Get that superheroie by the - girl.
  \item A woman in a black dress and glasses is on the news.
  \item A woman in a bikini is talking to a man.
  \item A bunch of bottles of liquor on a shelf.
  \item A close up of a camera with a pen on it.
  \item A person holding a white card with a black and white pattern.
  \item A doll is standing on a bed.
  \item Blur of a person walking.
  \item A group of people are gathered around a tree.
  \item A woman is sitting down on the news.
  \item A group of people walking around a street.
  \item A man in a blue shirt is standing next to a motorcycle.
  \item A person is putting a bag of food into a box.
  \item A person walking in the snow near a fence.
  \item A white plate with the words news brief on it.
  \item A man in a hat and a baseball cap.
  \item A white microwave oven.
  \item A white pot and a silver spoon on a table.
  \item A bunch of books on a table.
  \item The adobe file in adobe.
  \item A table with bowls of food and a bowl of food.
  \item A bowl filled with food sitting on top of a table.
  \item A bunch of plastic bags sitting on top of a table.
  \item Two dolls are sitting in a hospital bed.
  \item A flooded street in the suburbs of detroit, michigan.
  \item A small white mouse is sitting on the floor.
  \item A cat is sitting on the floor next to a bottle of liquid.
  \item A baseball player is being hit by a umpire.
  \item A cartoon character holding a white cat.
  \item A cat is sitting on the floor next to a bottle of liquid.
  \item A snow covered parking lot with a sign.
  \item A flooded street in phoenix, arizona.
\end{enumerate}
\subsection*{A.2 Multi Scene Prompts with format} 

\begin{itemize}
  \item[] \textbf{Example format:} \textit{previous scene: ...; then next scene: ...}
\end{itemize}

\begin{enumerate}
\item previous scene: a man and woman sitting at a table on the beach; then next scene: a woman sitting at a table with a drink
\item previous scene: a group of tents are set up in the woods; then next scene: a bird flying over the water at sunset
\item previous scene: a man and woman sitting at a table with drinks; then next scene: a woman in a bikini is standing on the beach
\item previous scene: a girl with long hair and green eyes stands in front of a tree; then next scene: a painting of a forest with trees and grass
\item previous scene: a boat is in the water near a rocky mountain; then next scene: a woman sitting at a table with a drink
\item previous scene: a yellow and black bird flying through a blue sky; then next scene: the girls of the twilight
\item previous scene: a little girl in a wheelchair with a toy; then next scene: a doll sitting in a chair next to a box
\item previous scene: a group of women holding signs in front of a crowd; then next scene: a man and woman are standing in front of a microphone
\item previous scene: a tall tower with a clock on top; then next scene: a man is putting his ballot in the ballot box
\item previous scene: a man in a suit and tie is talking to a woman; then next scene: a man in a suit and tie is talking to another man in a suit
\item previous scene: get that superheroie by the - girl; then next scene: file file for you png file for you my little pony
\item previous scene: a woman in a black dress and glasses is on the news; then next scene: a woman sitting on a couch in front of a tv screen
\item previous scene: a woman in a bikini is talking to a man; then next scene: a man and woman sitting at a table with drinks
\item previous scene: a bunch of bottles of liquor on a shelf; then next scene: a man is standing at the bar
\item previous scene: a close up of a camera with a pen on it; then next scene: a man standing in front of a motorcycle
\item previous scene: a person holding a white card with a black and white pattern; then next scene: a man is holding a cell phone
\item previous scene: a doll is standing on a bed; then next scene: a little girl is putting a gift box
\item previous scene: blur of a person walking; then next scene: a purple vase with a white flower on it
\item previous scene: a group of people are gathered around a tree; then next scene: a cat is standing in the dark
\item previous scene: a woman is sitting down on the news; then next scene: two women sitting on a couch talking to each other women
\item previous scene: a group of people walking around a street; then next scene: a woman walking down a street with a blue jacket
\item previous scene: a man in a blue shirt is standing next to a motorcycle; then next scene: a close up of a cell phone
\item previous scene: a person is putting a bag of food into a box; then next scene: a person is putting food into a container
\item previous scene: a person walking in the snow near a fence; then next scene: a black background with a white and red flower
\item previous scene: a white plate with the words news brief on it; then next scene: a woman standing in front of a brick wall
\item previous scene: a man in a hat and a baseball cap; then next scene: police investigates a man who was shot in the back of a car in the river
\item previous scene: a white microwave oven; then next scene: a white bowl with a spoon and a cup
\item previous scene: a white pot and a silver spoon on a table; then next scene: a white crocked pot
\item previous scene: a bunch of books on a table; then next scene: a table with a bunch of boxes of food
\item previous scene: the adobe file in adobe; then next scene: a computer screen with a green background
\item previous scene: a table with bowls of food and a bowl of food; then next scene: ingredients for making a cake
\item previous scene: a bowl filled with food sitting on top of a table; then next scene: a white cup with a spoon in it
\item previous scene: a bunch of plastic bags sitting on top of a table; then next scene: a pile of plastic bags
\item previous scene: two dolls are sitting in a hospital bed; then next scene: two dolls sitting on a chair
\item previous scene: a flooded street in the suburbs of detroit, michigan; then next scene: a dog is standing in the middle of a flooded street
\item previous scene: a small white mouse is sitting on the floor; then next scene: a small dog is sitting on the floor
\item previous scene: a cat is sitting on the floor next to a bottle of liquid; then next scene: a small white mouse
\item previous scene: a baseball player is being hit by a umpire; then next scene: a baseball player is about to catch the ball
\item previous scene: a cartoon character holding a white cat; then next scene: a cartoon character with a blue background
\item previous scene: a cat is sitting on the floor next to a bottle of liquid; then next scene: a cat is sitting on the floor next to a bottle of sauce
\item previous scene: a snow covered parking lot with a sign; then next scene: a black background with a white and red flower
\item previous scene: a flooded street in phoenix, arizona; then next scene: a police tape is taped around a wall that was covered with graffiti
\end{enumerate}

\subsection*{B.3 Multi Scene Prompts without format} 

\begin{enumerate}
  \item A man and woman sitting at a table on the beach; a woman sitting at a table with a drink.
  \item A group of tents are set up in the woods; a bird flying over the water at sunset.
  \item A man and woman sitting at a table with drinks; a woman in a bikini is standing on the beach.
  \item A girl with long hair and green eyes stands in front of a tree; a painting of a forest with trees and grass.
  \item A boat is in the water near a rocky mountain; a woman sitting at a table with a drink.
  \item A yellow and black bird flying through a blue sky; the girls of the twilight.
  \item A little girl in a wheelchair with a toy; a doll sitting in a chair next to a box.
  \item A group of women holding signs in front of a crowd; a man and woman are standing in front of a microphone.
  \item A tall tower with a clock on top; a man is putting his ballot in the ballot box.
  \item A man in a suit and tie is talking to a woman; a man in a suit and tie is talking to another man in a suit.
  \item Get that superheroie by the - girl; file file for you png file for you my little pony.
  \item A woman in a black dress and glasses is on the news; a woman sitting on a couch in front of a tv screen.
  \item A woman in a bikini is talking to a man; a man and woman sitting at a table with drinks.
  \item A bunch of bottles of liquor on a shelf; a man is standing at the bar.
  \item A close up of a camera with a pen on it; a man standing in front of a motorcycle.
  \item A person holding a white card with a black and white pattern; a man is holding a cell phone.
  \item A doll is standing on a bed; a little girl is putting a gift box.
  \item Blur of a person walking; a purple vase with a white flower on it.
  \item A group of people are gathered around a tree; a cat is standing in the dark.
  \item A woman is sitting down on the news; two women sitting on a couch talking to each other women.
  \item A group of people walking around a street; a woman walking down a street with a blue jacket.
  \item A man in a blue shirt is standing next to a motorcycle; a close up of a cell phone.
  \item A person is putting a bag of food into a box; a person is putting food into a container.
  \item A person walking in the snow near a fence; a black background with a white and red flower.
  \item A white plate with the words news brief on it; a woman standing in front of a brick wall.
  \item A man in a hat and a baseball cap; police investigates a man who was shot in the back of a car in the river.
  \item A white microwave oven; a white bowl with a spoon and a cup.
  \item A white pot and a silver spoon on a table; a white crocked pot.
  \item A bunch of books on a table; a table with a bunch of boxes of food.
  \item The adobe file in adobe; a computer screen with a green background.
  \item A table with bowls of food and a bowl of food; ingredients for making a cake.
  \item A bowl filled with food sitting on top of a table; a white cup with a spoon in it.
  \item A bunch of plastic bags sitting on top of a table; a pile of plastic bags.
  \item Two dolls are sitting in a hospital bed; two dolls sitting on a chair.
  \item A flooded street in the suburbs of detroit, michigan; a dog is standing in the middle of a flooded street.
  \item A small white mouse is sitting on the floor; a small dog is sitting on the floor.
  \item A cat is sitting on the floor next to a bottle of liquid; a small white mouse.
  \item A baseball player is being hit by a umpire; a baseball player is about to catch the ball.
  \item A cartoon character holding a white cat; a cartoon character with a blue background.
  \item A cat is sitting on the floor next to a bottle of liquid; a cat is sitting on the floor next to a bottle of sauce.
  \item A snow covered parking lot with a sign; a black background with a white and red flower.
  \item A flooded street in phoenix, arizona; a police tape is taped around a wall that was covered with graffiti.
\end{enumerate}

\clearpage
\appendix
\section*{Appendix C: Video Transition Clip Extraction Code}

\begin{figure}[h!]
\centering
\begin{tabular}{|p{\textwidth}|}
\hline
\textbf{Python Code for Scene Transition Detection and Clip Extraction:} \\
\texttt{import json} \\
\texttt{import cv2} \\
\texttt{import numpy as np} \\
\texttt{import pandas as pd} \\
\texttt{import ffmpeg} \\
\texttt{from scenedetect import detect, ContentDetector} \\
\texttt{from tqdm import tqdm} \\
\texttt{import os} \\\\

\texttt{\# Configuration parameters} \\
\texttt{CLIP\_LENGTH = 10 \# target duration in seconds} \\
\texttt{PADDING = 5 \# padding before and after transition point} \\
\texttt{MIN\_SCENE\_LENGTH = 3} \\
\texttt{MAX\_SCENE\_LENGTH = 10} \\\\

\texttt{def detect\_scenes(video\_path):} \\
~~~\texttt{"Detect scene transitions using PySceneDetect"} \\
~~~\texttt{scene\_list = detect(video\_path, ContentDetector())} \\
~~~\texttt{return [scene[1].get\_seconds() for scene in scene\_list]} \\\\

\texttt{def extract\_transitional\_clips(video\_path, scene\_timestamps):} \\
~~~\texttt{video\_name = os.path.basename(video\_path).split('.')[0]} \\
~~~\texttt{output\_clips = []} \\
~~~\texttt{cap = cv2.VideoCapture(video\_path)} \\
~~~\texttt{fps = cap.get(cv2.CAP\_PROP\_FPS)} \\
~~~\texttt{total\_frames = int(cap.get(cv2.CAP\_PROP\_FRAME\_COUNT))} \\
~~~\texttt{video\_duration = total\_frames / fps} \\
~~~\texttt{for timestamp in scene\_timestamps:} \\
~~~~~~\texttt{start\_time = max(0, timestamp - PADDING)} \\
~~~~~~\texttt{end\_time = min(video\_duration, timestamp + PADDING)} \\
~~~~~~\texttt{if end\_time - start\_time > MAX\_SCENE\_LENGTH:} \\
~~~~~~~~~~~\texttt{end\_time = start\_time + MAX\_SCENE\_LENGTH} \\
~~~~~~\texttt{output\_filename = f"\{video\_name\}\_\{int(start\_time)\}-\{int(end\_time)\}.mp4"} \\
~~~~~~\texttt{output\_path = os.path.join(OUTPUT\_VIDEO\_DIR, output\_filename)} \\
~~~~~~\texttt{ffmpeg.input(video\_path, ss=start\_time, to=end\_time)} \\
~~~~~~~~~~~\texttt{.output(output\_path, vcodec="libx264", acodec="aac")} \\
~~~~~~~~~~~\texttt{.run(overwrite\_output=True, quiet=True)} \\
~~~~~~\texttt{output\_clips.append(\{} \\
~~~~~~~~~~~\texttt{"file\_path": output\_path, "video\_name": video\_name,} \\
~~~~~~~~~~~\texttt{"start\_time": start\_time, "end\_time": end\_time,} \\
~~~~~~~~~~~\texttt{"duration": end\_time - start\_time, "transition\_frame": timestamp} \\
~~~~~~\texttt{\})} \\
~~~\texttt{cap.release()} \\
~~~\texttt{return output\_clips} \\\\

\hline
\end{tabular}
\caption{Python code for detecting scene transitions and extracting fixed-length video clips centered on transitions.}
\label{fig:scene_clip_code}
\end{figure}

\begin{figure}[h!]
\centering
\begin{tabular}{|p{\textwidth}|}
\hline

\texttt{def validate\_clips(clips):} \\
~~~\texttt{filtered\_clips = []} \\
~~~\texttt{for clip in tqdm(clips, desc="Validating Clips"):} \\
~~~~~~\texttt{cap = cv2.VideoCapture(clip["file\_path"])} \\
~~~~~~\texttt{prev\_frame = None; transition\_detected = False} \\
~~~~~~\texttt{while cap.isOpened():} \\
~~~~~~~~~~~\texttt{ret, frame = cap.read(); if not ret: break} \\
~~~~~~~~~~~\texttt{if prev\_frame is not None:} \\
~~~~~~~~~~~~~~\texttt{diff = np.mean(cv2.absdiff(prev\_frame, frame))} \\
~~~~~~~~~~~~~~\texttt{if diff > 50: transition\_detected = True; break} \\
~~~~~~~~~~~\texttt{prev\_frame = frame} \\
~~~~~~\texttt{cap.release()} \\
~~~~~~\texttt{if transition\_detected: filtered\_clips.append(clip)} \\
~~~\texttt{return filtered\_clips} \\\\

\texttt{def save\_metadata\_to\_json(filtered\_clips):} \\
~~~\texttt{output\_data = [\{"file\_path": c["file\_path"], "text": ""\} for c in filtered\_clips]} \\
~~~\texttt{with open(OUTPUT\_JSON\_FILE, "w", encoding="utf-8") as f:} \\
~~~~~~\texttt{json.dump(output\_data, f, ensure\_ascii=False, indent=4)} \\
~~~\texttt{print(f"Metadata saved to \{OUTPUT\_JSON\_FILE\}")} \\\\

\texttt{def main():} \\
~~~\texttt{video\_path = ".../input\_videos/example.mp4"} \\
~~~\texttt{scene\_timestamps = detect\_scenes(video\_path)} \\
~~~\texttt{video\_clips = extract\_transitional\_clips(video\_path, scene\_timestamps)} \\
~~~\texttt{validated\_clips = validate\_clips(video\_clips)} \\
~~~\texttt{save\_metadata\_to\_json(validated\_clips)} \\\\

\texttt{if \_\_name\_\_ == "\_\_main\_\_": main()} \\

\hline
\end{tabular}
\caption{Python code for detecting scene transitions and extracting fixed-length video clips centered on transitions.(continued)}
\label{fig:scene_clip_code}
\end{figure}

\end{document}